\documentclass[11pt]{article}
\usepackage{acl2014}
\usepackage[utf8]{inputenc}
\usepackage{times}
\usepackage{url}
\usepackage{latexsym}
\usepackage{amsmath}
\usepackage{paralist}
\usepackage{pdfpages}
\DeclareMathSizes{11}{9}{13}{9}

\title{Empirical Evaluation of Tree distances for Parser Evaluation}
\author{Taraka Rama K.\\
Språkbanken\\
University of Göteborg}

\date{}

\begin{document}
\maketitle
\begin{abstract}
In this empirical study, I compare various tree distance measures -- originally 
developed in computational biology for the purpose of tree comparison -- for 
the purpose of parser evaluation. I will control for the parser setting by 
comparing the automatically generated parse trees from the state-of-the-art 
parser~\cite{Charniak:00} with the gold-standard parse trees. The article 
describes two different tree distance measures (RF and QD) along with its 
variants (GRF and GQD) for the purpose of parser evaluation. The article will 
argue that RF measure captures similar information as the standard EvalB 
metric~\cite{sekine1997evalb} and the tree edit distance~\cite{zhang1989simple} 
applied by \newcite{tsarfaty2011evaluating}. Finally, the article also provides 
empirical evidence by reporting high correlations between the different tree 
distances and EvalB metric's scores.
\end{abstract}

\section{Introduction}
The \emph{tree}, as a device, has been employed to depict the relationships 
both within languages as well as species. Both historical linguists and 
evolutionary biologists employ the tree device to capture language and 
biological evaluation. The computational subbranches of both the disciplines 
employ statistical and quantitative techniques to infer relationships based on 
sequence data: linguistic units such as lexemes, grammatical markers; and gene 
sequences~\cite{levinson2012tools,felsenstein2004inferring}. In the case of 
historical linguistics, inferred language trees are compared with trees 
inferred from application of the comparative 
method~\cite{hoenigswald1973comparative}. \newcite{Wichmann:2011:2210-5824:205} 
and \newcite{pompei2011accuracy} compare different tree distance measures for 
comparing different language relationship inference techniques.

In the related field of computational linguistics, the grammatical 
relationships between words in a sentence are captured through rooted, labeled 
trees. Moving along the spectrum, trees by themselves play a major role in 
computer science~\cite{bille2005survey}. \newcite{goodman1996parsing} formalizes 
the PARSEVAL metrics \cite{abney1991procedure} in the context of 
constituency-based parsing. In any case, much research seems to be have 
gone into quantitative tree comparison in related disciplines.

The rest of the article is organized as followed: Section~\ref{sec:treedist} 
describes tree distance measures. Section~\ref{sec:dataset} describes the 
dataset used in our experiments. In section~\ref{sec:results}, I describe the 
results of the empirical study. Section~\ref{sec:concl} concludes and provides 
the future directions.

\section{Tree distances}\label{sec:treedist}
\subsection{Basics}\label{subsec:basics}
In this subsection, I define basic symbols and terms. A input string $I$ is 
composed of words $w_1\cdots w_n$ of length $n$. A parse tree, $T$ defines the 
relationship between these words. Following \newcite{goodman1996parsing}, a 
tree $T$ is composed of triples $(s,t,X)$ where, $s$ to $t$ are consecutive 
words dominated by an internal node labeled as $X$. Following 
\newcite{goodman1996parsing}, the matching criteria for an automatic parse tree 
$T_a$ with its gold standard tree $T_g$ is defined in terms of unlabeled 
precision ($P$) and recall ($R$). Let $N_g$ be the number of internal nodes in 
$T_g$ and $N_a$ be the number of internal nodes in $T_a$. Let $B = |\{(s,t,X): 
(s,t,X) \in T_a \land (s,t,Y) \in T_g\}|$.

\begin{itemize}
 \item Precision ($P$): $B/N_a$
\item Recall ($R$): $B/N_g$
\item F-score: $2PR/(P+R)$
\end{itemize}

Let $E$ be the set of edges, $V$ be the set of internal nodes, $IE$ be 
the set of internal edges and, $n$ the number of leaves in a tree. Then 
the following conditions hold for any tree:
\begin{itemize}
 \item $|E| = |V| + n-1$
\item $|IE|=|V|-1$
\item $|V| = n-2$
\end{itemize}
The rest of the article assumes that trees are both unlabeled and $m$-ary.

\subsection{RF (Robinson-Foulds) Distance}\label{subsec:bipart}
\newcite{robinson1981comparison} defined two operations $\alpha$ and 
$\alpha^{-1}$ for transforming a tree $T$ to tree $T'$ in a finite sequence of 
operations. The $\alpha$ operation is a edge contraction operation whereas 
$\alpha^{-1}$ is an expansion operation. It takes a maximum of $|E|+|E'|$ 
operations and a minimum of $||E|-|E'||$ operations to transform $T$ to $T'$. 
In essence, the RF distance is defined as the number of $\alpha^{-1}$ and 
$\alpha$ operations needed to transform one tree to another. An internal edge 
divides a tree into two disjoint sets of leaf nodes known as bipartition. A 
tree with $n$ leaves has $n-3$ bipartitions. Consequently, an internal edge 
defines a bipartition. In terms of bipartitions, RF distance, $RFD$ between two 
trees $T, T'$ is defined as:
\begin{itemize}
 \item $\frac{|E-E'|+|E'-E|}{|E|+|E'|}$
\item $1-\frac{2|E\cap E'|}{|E|+|E'|}$
\end{itemize}

Thus, the RF distance measures the dissimilarity in the topology between the 
inferred tree and the corresponding family tree. It should be noted that the
RF distance does not take branch lengths into account. Also, the RF distance 
can be further modified to compute the distance between two trees by 
introducing a label substitution operation. It can be easily seen that RF 
distance is related to tree edit distance used in 
\newcite{tsarfaty2011evaluating}. RF distance works in terms of internal edges 
whereas, tree edit distance works in terms of node insertion, deletion, and 
substitution. I also show that $P$ and $R$ are related to $RFD$.
\begin{itemize}
 \item $P = |E_a\cap E_g|/|E_a|$
\item $R = |E_a\cap E_g|/|E_g|$
\item Substituting, $P$ and $R$ into the formulas for $RFD$ yields another 
formula: $1-\frac{2RP}{R+P}$.
\end{itemize}
This formulation suggests that the F-score, of each parse tree, obtained from 
the EvalB metric should correlate highly with the RF distance. The RF distance 
is zero when both the trees have no internal edges. Such trees have star 
topologies.

RF distance is a harsh measure that penalizes $T_a$ for small errors. 
For example, a triple $(s,t_{-1},X)$ -- assuming that position of $s$ is not 
the same as $t$ -- in $T_a$ will be counted as an error since it is not present 
as $(s,t,X)$ in $T_g$. Such kind of harsh penalization can be smoothened by 
employing Generalized Robinson-Foulds distance (GRFD). 

GRFD, as introduced by \newcite{pompei2011accuracy}, relaxes the strict 
equality condition of bipartitions with \emph{compatibility} criterion. To 
start with, a bipartition $k$ divides the leaf set into two disjoint sets 
$k_1,k_2$. In the authors' terms, a bipartition $a$ in tree $T_a$ is said to be 
compatible with $i$-th bipartition in $T_g$ -- consisting of $g_1^i, g_2^i$ sets 
-- if for each bipartition $i$ in $T_g$, one of the following is true: 
$a_1\subseteq g_1^i$, or $a_1\subseteq g_2^i$, or $a_2\subseteq g_1^i$, or 
$a_2\subseteq g_2^i$. Finally, GRFD is defined as 
$\frac{|E_a|-|C(T_a,T_g)|}{|E_a|}$ where, $C(T,T')$ yields a set of compatible 
bipartitions. The GRFD is not a metric whereas, RFD is shown to 
be a metric~\cite{robinson1981comparison}.\footnote{I use my own implementation 
to compute both RFD and GRFD.}

\subsection{Quartet distance}\label{subsec:qdist}
Both RFD and GRFD work with internal edges (or subtrees). Another possible 
distance is the quartet distance which measures the distance between two trees 
in terms of the number of different quartets between the two 
trees~\cite{estabrook1985comparison}. A quartet is defined as a set of four 
leaves selected from a set of leaves without replacement. A tree with $n$ 
leaves has ${n \choose 4}$ quartets in total.

A quartet defined on four leaves $a,b,c,d$ can have four different topologies: 
$ab|cd$, $ac|bd$, $ad|bc$, and $ab\times cd$. The first three topologies have 
an internal edge separating two pairs of leaves. Such quartets are called as 
\emph{butterflies}. The fourth quartet has no internal edge and as such is 
known as star quartet. A parse tree can have an internal node that is a parent 
to at least four leaves. By definition, quartet distance is defined only for 
those sentences whose length is at least $4$.

For a tree $T$ with $n$ leaves, the quartets can be partitioned into 
set of butterflies, $B(T)$, and set of stars, $S(T)$. Then, quartet distance 
(QD) between $T$ and $T'$ is defined as:
\begin{equation}
1-\frac{|S(T)\cap S(T')|+|B(T)\cap 
B(T')|}{{n \choose 4}} 
\end{equation}

\newcite{christiansen2006fast} reformulate QD as 
follows:
\begin{equation}
\label{eq:qd}
\frac{ B(T)+B(T')-2|B(T)\cap B(T')|- DB(T, T')}{{n
\choose 4}}
\end{equation}
\normalsize
where, $DB(T, T')$ is the number of different butterflies between $T, T'$. A 
different butterfly is based on the same leaf set but has different topologies 
in the two trees. Since the trees are $m$-ary, $d$ is defined as the maximum 
degree of an internal node. \newcite{christiansen2006fast} developed a fast 
algorithm that runs in $\mathcal{O}(n^2d^2)$ in time, and needs 
$\mathcal{O}(n^2)$ in terms of space.\footnote{Available at 
\url{http://birc.au.dk/software/qdist/}.}

The QD formula in equation \ref{eq:qd} counts the butterflies in $T_a$ as 
errors. The tree $T_a$ should not be penalized for the unresolvedness in the 
gold standard tree $T_g$. To this end, \newcite{pompei2011accuracy} defined a 
new measure known as GQD (Generalized QD) to discount the presence of star 
quartets in $T_g$. GQD is defined as $DB(T_a, T_g)/B(T_g)$. We use
both QD and GQD to evaluate the quality of the automatically generated parse 
trees. It can be easily seen that both QD and GQD are lenient in scoring a 
subtree when the parser misses or adds a leaf in comparison with the gold 
standard tree.

\section{Parser and dataset}\label{sec:dataset}
For our experiments, I use the Charniak parser to parse the $2416$ sentences 
in the section $23$ of the Penn TreeBank. Out of these parses, I discarded all 
the parses which have sentence length less than $4$. I was left with $2378$ 
parses after this step. I also converted both the automatic and gold parses 
into the NEWICK format~\cite{felsenstein2004inferring}. NEWICK format is a 
bracketing format to represent trees and uses ``,'' symbol to separate 
adjacent leaves; opening and closing parantheses. I also removed the POS tag 
for each word. This step removes all the nodes which have a degree of $2$. Hence 
a final NEWICK format tree consists of brackets and words alone. I use the 
EvalB program to compute the sentence-level precision and recall scores.

\section{Results}\label{sec:results}
In this paper, I only work with sentence-level parsing F-scores. Our first 
hypothesis is that RFD will correlate to a large extent with the F-scores. This 
hypothesis is formed from the similarity in the formulae shown in 
section~\ref{subsec:bipart}. Accordingly, the pearson's $r$ between RFDs and 
F-scores is $-0.9$. We attribute the difference to the handling of top-level 
brackets. We ignore the top-level bracket enclosing the root symbol whereas 
EvalB includes it in the calculation. GRFDs and F-scores also show a 
correlation of $-0.82$. The correlation is expected to be negative since 
F-score is a measure of similarity whereas RFD is a distance metric.

I also checked the correlation of QDs and GQDs with F-scores. The correlation 
is quite high: $-0.73$ and $-0.68$ respectively. I also computed the 
correlation between the four different distance measures. The correlation is 
shown in table~\ref{tab:corrs}.

\begin{table}[h]
\centering
\small
\begin{tabular}{|l|cc|}
\hline
& QD & GQD \\\hline
RFD & 0.86 & 0.79 \\
GRFD & 0.78 & 0.89 \\
\hline
\end{tabular}
\caption{Correlations between all the distance measures.}
\label{tab:corrs}
\end{table}

The measures correlate highly but not completely. This is in line with the 
observation of \newcite{christiansen2006randers} who observe -- based on 
simulated trees and real world data -- that RFD and QD correlate quite well but 
measure different aspects of the trees.

\section{Conclusion and future work}\label{sec:concl}
In this article, I described two popular tree distance measures from 
computational biology and applied them for the purpose of parser evaluation. I 
observe that the measures correlate with each other, to a large extent, for a 
single parser's output. I argued that QD and generalized tree distance measures 
are much suitable for parser evaluation task since the gold parses can have 
internal nodes with degrees greater than $2$. I also argued that RF distance 
measures the same thing as the tree edit distance does for computing unlabeled 
accuracies. This hypothesis is corroborated in terms of a high pearson's $r$. 
In future, we plan to test the different measures on other off-the-shelf 
parsers.

\section*{Acknowledgements}
I thank Richard Johansson for providing the binaries of Charniak parser. I 
thank Prasanth Kolachina and Richard Johansson for interesting discussions. I 
thank Joakim Nivre for the patience and 
encouragement through the whole process of experiments and report writing.

\bibliographystyle{acl}
\bibliography{myreflnks}

\end{document}